\documentclass[letterpaper, 10 pt, conference]{ieeeconf}
\IEEEoverridecommandlockouts
\let\labelindent\relax
\usepackage{enumitem}

\usepackage[utf8]{inputenc}
\usepackage{hyperref}
\usepackage{graphicx}
\usepackage{xcolor}
\usepackage[font=small,labelfont=bf]{caption}
\usepackage{subcaption}

\usepackage{acronym}
\acrodef{RPPU}{Relative Predictive Proxy Usefulness}
\acrodef{PPU}{Predictive Proxy Usefulness}
\acrodef{PPV}{Proxy Predictivity Value}
\acrodef{PRPV}{Proxy Relative Predictivity Value}
\acrodef{PLV}{Proxy Learning Value}
\acrodef{AIDO}{AI Driving Olympics}
\acrodef{SRCC}{Sim-vs-Real Correlation Coefficient}
\acrodef{POD}{Proxy Observation Discrepancy}

\usepackage{amsmath}

\usepackage{amsthm}
\usepackage{amssymb}
\usepackage[noadjust]{cite}

\graphicspath{ {./images/} }

\makeatletter
\theoremstyle{definition}
\makeatother

\usepackage{babel}

\newtheorem{definition}{Def.}

\title{On Assessing the Usefulness of Proxy Domains for Developing and Evaluating Embodied Agents}
\author{Anthony Courchesne,${}^{1,*}$ Andrea Censi,${}^{2}$ Liam Paull${}^{1}$
\thanks{${}^1$ Montréal Robotics and Embodied AI Lab (REAL) at Université de Montéal, Mila.;
${}^2$\,ETH Z\"urich, Switzerland;
${}^*$\,Corresponding author anthony.courchesne@mail.mcgill.ca}}
\date{\today}

\begin{document}

\maketitle

\begin{abstract}

In many situations it is either impossible or impractical to develop and evaluate agents entirely on the target domain on which they will be deployed. This is particularly true in robotics, where doing experiments on hardware is much more arduous than in simulation. This has become arguably more so in the case of learning-based agents. To this end, considerable recent effort has been devoted to developing increasingly realistic and higher fidelity simulators. However, we lack any principled way to evaluate how good a ``proxy domain'' is, specifically in terms of how \textit{useful} it is in helping us achieve our end objective of building an agent that performs well in the target domain. In this work, we investigate methods to address this need. 
We begin by clearly separating two uses of proxy domains that are often conflated: 1) their ability to be a faithful predictor of agent performance and 2) their ability to be a useful tool for learning. In this paper, we attempt to clarify the role of proxy domains and establish new \textit{proxy usefulness} (PU) metrics to compare the usefulness of different proxy domains. We propose the \textit{relative predictive PU} to assess the predictive ability of a proxy domain and the \textit{learning PU} to quantify the usefulness of a proxy as a tool to generate learning data. Furthermore, we argue that the value of a proxy is conditioned on the \textit{task} that it is being used to help solve. We demonstrate how these new metrics can be used to optimize parameters of the proxy domain for which obtaining ground truth via system identification is not trivial.
\end{abstract}

\section{Introduction}


Developing and evaluating agents for physically embodied systems such as robots in the setting that they are meant to be deployed in can be costly, dangerous, and time consuming. As a result, it is often desirable to have some \textit{proxy} of the target task domain, which can be a simulation environment or some simpler smaller scale real environment. The fidelity or accuracy of this proxy domain is important in the sense that we need it to be a \textit{faithful predictor} of performance on the target domain.  

Recent breakthroughs in machine learning have increased the popularity of data-driven approaches to solve tasks with embodied agents. Two prevalent paradigms include reinforcement \cite{mataric1997reinforcement} and imitation \cite{finn2017one} learning. As a result, the value of data has increased dramatically. Obtaining data from robots is costly since it requires deployment of hardware, operation of the robot and time-consuming rollouts of policies. 
Moreover, the policy exploration process might end up in hardware damage while unsafe states are explored. 
This further motivates the need for a proxy domain, but the objectives here are somewhat different.  
In this case we want to \textit{efficiently} acquire knowledge in the proxy domain that can be \textit{transferred} to the target domain. 

There may be other advantages of the proxy domain for the purposes of agent learning, such as: the availability of ground-truth data such as robot localization, reduced cost of data collection,  ability to generate and randomize large numbers of samples, customization of scenarios to facilitate the learning of edge cases, parallelization of the data generation process, and others.

\begin{figure}[t]
\centering
\includegraphics[width=\columnwidth]{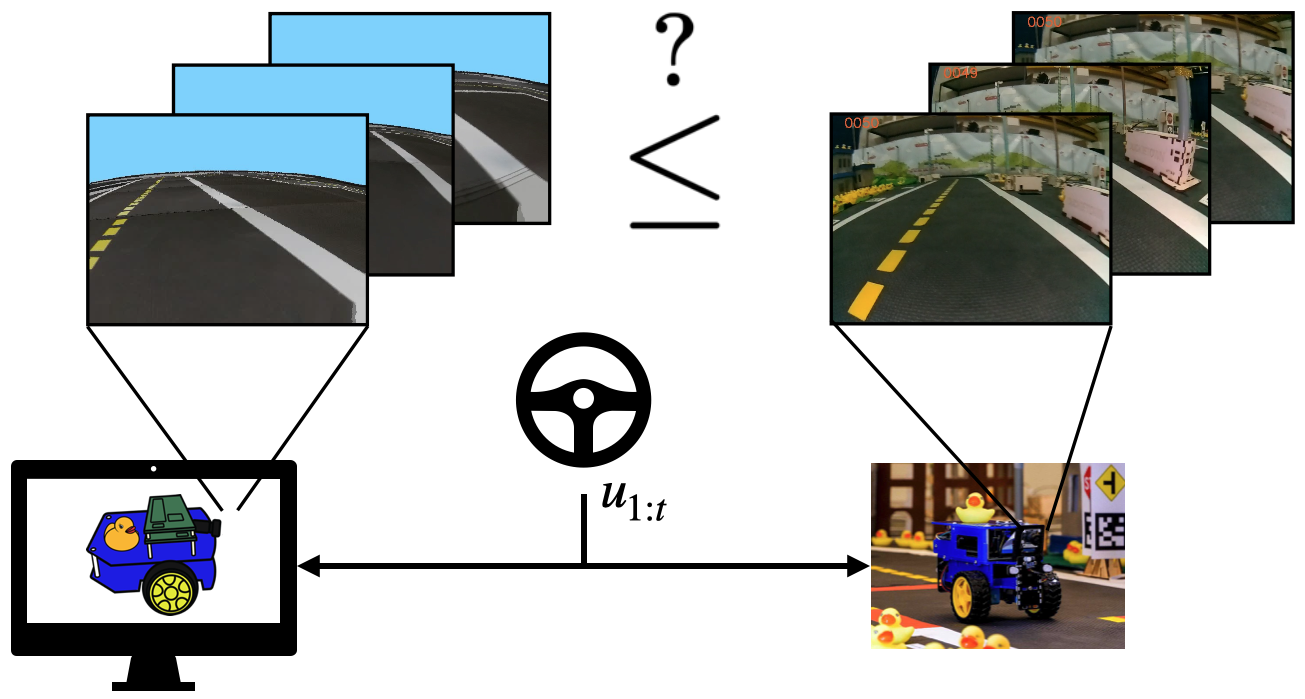}
\caption{How do we compare the value of two different instantiations of the same task? This is a common setting in robotics where it is easier to develop and evaluate on a simulator or proxy domain before deploying on real hardware (the target domain). If we start the agent at the same state in both domains and execute the same sequence of control signals, we will get different output measurements due to discrepancies in dynamics, appearance, rendering, and many other factors. We argue that directly comparing these observations is not an appropriate measure of the proxy's value, since it does not consider the consequence of the actions under the task specification. 
Instead, we propose measures for assessing the value quantitatively both in terms of how useful the proxy domain is in terms of predictivity and as a tool for learning agents. }
\label{fig:splash}
\end{figure}

Any domain used as a proxy for a target domain introduces a set of discrepancies, often resulting in performance reduction. The ``reality gap'' is a term that is typically used in the context of sim-to-real, although the concept can be generalized to any domain transfer. For this reason, we will instead use the term ``domain gap'' as a more generic case.
The domain gap has been observed in multiple works, and the community has found multiple ways to mitigate its effects (i.e. "crossing the gap"), without actually explicitly quantifying it. As a result, algorithms that show good performance in some settings (e.g. when the domain discrepancy is small) may completely fail in others, and there are no methods of predicting transferability a priori.


In this work, we quantify the usefulness of a proxy domain by providing \textit{proxy usefulness} (semi)metrics.
We make a clear distinction between proxies used to predict task performance (used to select agents according to their performance) and data-generating proxies (used to generate data samples or policies). In the first case, we propose the Proxy Relative Predictivity Value (PRPV), a metric to quantify the predictivity of a proxy, enabling researchers to find the most predictive proxy available to them. 
We also prescribe a metric to assess the usefulness of a proxy to generate learning data or trained agents, giving the possibility to researchers to compare different data-generating domains and select the one that yields the best agents.

Our new metrics allow robotics practitioners to tune some parameters of their proxy domain for which system identification is non-trivial and ground truth value for the target domain is often not available, such as observation blur, input delay, field of view, etc.


%
%

\section{Related work}

In the most common configuration, the source domain is a simulator and the target domain is a real robot, in which case the policy transfer is called "sim-to-real". However, it is also frequent to see "sim-to-sim" (\cite{james2019sim,golemo2018sim,yu2018policy}), or even "real-to-real", such as in the case of \cite{paull2017duckietown} where small, cheap robots are used to represent an expensive car, or in \cite{xia2019gibson} where authors used real-to-real to validate their sim-to-real performance. Our analysis of the domain discrepancies applies to all of these cases.

\subsection{Crossing the Domain Gap}
Multiple techniques exist to mitigate the effects of the domain gap. One such promising technique that has received a lot of attention lately is domain randomization \cite{tobin2017domain}: during training, a different proxy is sampled from a domain family (see \ref{sec:prelim}) for each run, preventing the model from overfitting to a specific instance.
In the best-case scenario, the target domain will appear as another sample from the same distribution. Data augmentation is another popular solution \cite{shrivastava2017learning, pashevich2019learning, bousmalis2017unsupervised} where a model is trained to augment data generated by a proxy to match that of the target domain.
Meta-learning and domain adaptation are also often used to transfer the knowledge acquired in a source domain to a different domain \cite{yu2018one, bousmalis2018using}.
The authors in \cite{mouret2013crossing, mouret201720} show that it is possible to find which controllers are most likely to transfer well to the target setup using only a few sample experiments on the real robot, a method they call ``The transferability approach''.
Crossing the domain gap was also shown to be possible by learning a common representation for both source and target domains, either by imposing a common state representation using an autoencoder and a disciminator network \cite{zhang2019adversarial} or by forming weakly aligned pairs of source and target data, effectively transferring the annotations to the target images \cite{tzeng2020adapting}.

\subsection{Optimizing Proxies}
In our work, we are more interested in the different ways to analyze the domain gap and improve the proxy rather than circumvent its effects. The first step to reduce the gap imposed by a proxy is to do proper system identification \cite{tan2018sim}. 
For a specific domain pair, it is possible to carry out standardized tests to find inconsistencies between models \cite{pepper2007robot}.
It is common \cite{chebotar2019closing,collins2020traversing,tan2016simulation} to use a task performance metric to optimize  proxy parameters such as actuator gains, masses, static friction, etc. Recent papers have also shown success in improving their proxies by adding a module that learns inverse dynamics to adapt the action that the proxy outputs to the expected action in the target domain \cite{christiano2016transfer,hanna2017grounded}.

\begin{figure}[tb]
\centering
\includegraphics[width=\columnwidth]{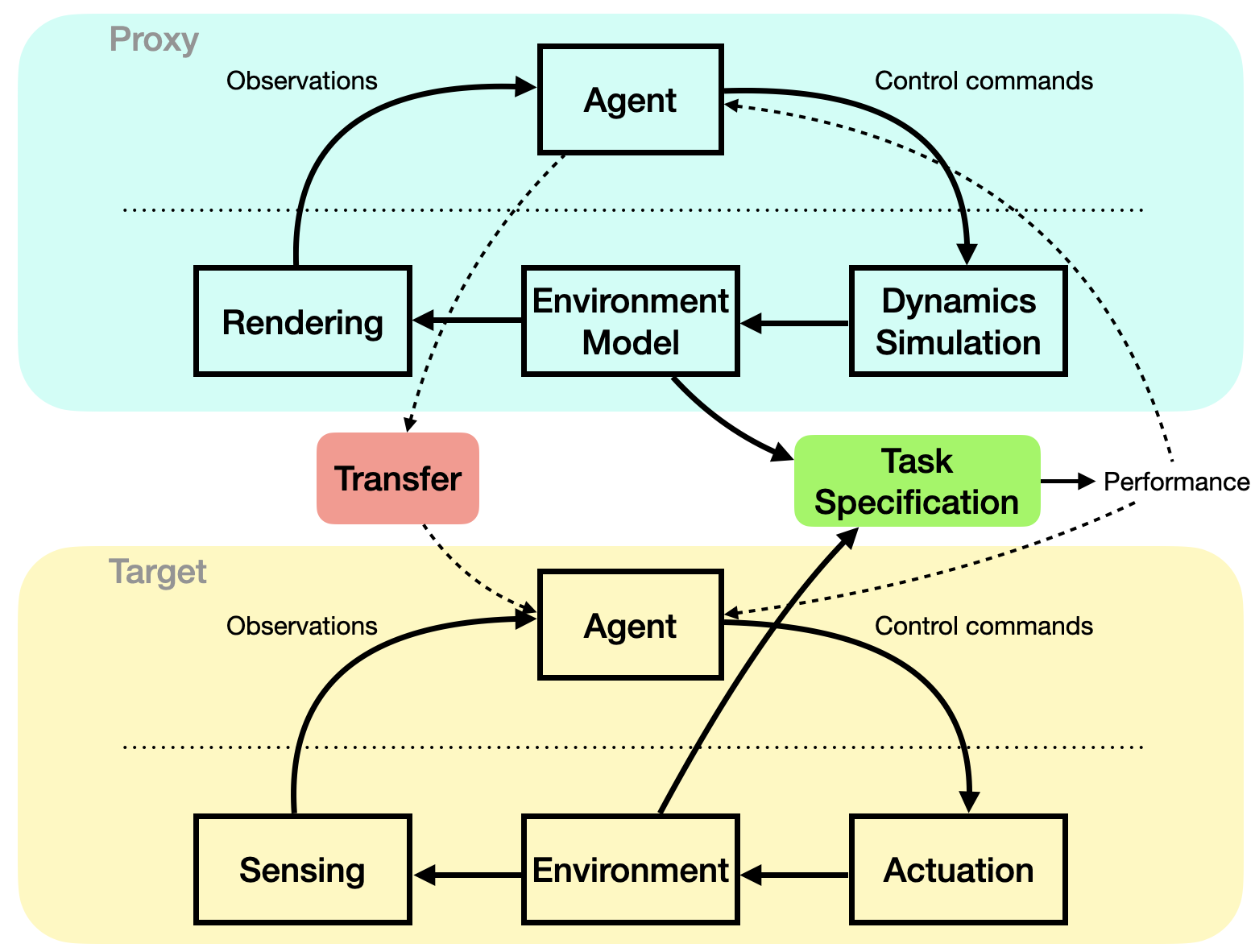}
\caption{The agent/environment interface for proxy (simulation) and the target (real robot) domains. An agent receives observations and generates control commands. The \textbf{predictive} value of a simulator lies in its ability to faithfully reproduce an estimate of the task performance. A simulator may also be used in a \textbf{learning} paradigm. In this case, the value of the simulator lies in how many fewer trials we need to perform on the real robot to achieve equivalent performance.}
\label{fig:agent-environment}
\end{figure}

\subsection{Quantifying Domain Discrepancies}
Being able to quantify the domain gap may lead to better prediction of the transferability of agents. As such, multiple attempts have been made by the community to obtain a metric to quantify this gap. One promising work in that direction establishes the sim-to-real disparity: a metric specific to an agent that predicts how well it will transfer to the real robot \cite{koos2012transferability}. By conducting only a few experiments both on the real robot and in simulation, it is possible to learn a surrogate model that will estimate the fitness function on the real robot via interpolation. However, it is assumed that we have access to a quick way to compute the distance between two controllers in a domain such that similar controllers will have a similar transferability (e.g. behavioral features), which is often not the case. Moreover, we are interested in a way to quantify the domain gap at the level of a simulator instead of at the agent level. We believe that the domain gap should be agnostic to the agent.

The ``$\nu$-gap'' \cite{vinnicombe1993frequency} is a metric used in robust control theory to analyze the discrepancy between two feedback control system. It has been used \cite{sorocky2020experience} 
to quantify the discrepancy between robotic systems without any knowledge about their dynamics since it compares two different controllers in a black-box fashion. The metric is computed by measuring the largest chordal distance between the two controllers projected onto the Riemann sphere. It is then possible to minimize it to find the most representative domain. This metric has the limitation of comparing only the dynamics and being only available for linear systems.

A version of ``simulator fidelity'' has been used as a metric in a multi-fidelity simulator environment \cite{cutler2014reinforcement} to define the fidelity of a domain $\Sigma_i$ to a target domain $\Sigma_j$. The fidelity is defined by the maximum error in the optimal value function. Using this metric, it is possible to optimize the training of reinforcement-learning-based agents using multiple simulators, taking into account the cost of generating data for high-fidelity simulators. On the other hand, this definition is inconvenient when used to assess a simulator since the value function is not always trivially computed, especially in the real world. We believe that the definition of the domain gap should not be tightly coupled with machine learning, and it should be independent of the algorithm used.

In \cite{kadian2020sim2real}, the authors were able to create a high-fidelity virtual domain for their task using real-to-sim. Given agents and a task, they used the Pearson correlation coefficient between runs on the real robot and in simulation to evaluate the reality gap. This allowed them to find that the noise model for their dynamics was wrong and that learning-based agents were overfitting to the simulator domain by using glitches. In their work, they frequently refer to rank inversion to validate their empirical results, which is an idea that is also used in \cite{ligot2018mimicking}. Like the former, we support that the predictive ability of a simulator lay in its capacity to predict the performance ordering of the agents. In our work, we further define the domain gap based on rank inversion. Using a relative metric for the reality gap allows us to use multiple performance metrics to get a domain gap measure through partial ordering. We additionally propose a separate and orthogonal measure of a simulator's value for learning. 



\section{Preliminaries}
\label{sec:prelim}

We will consider a domain to encapsulate the actuation, environment and sensing, as shown in Fig.~\ref{fig:agent-environment}. For a simulated domain, those components translate to a simulation of the dynamics, an environment model and rendering, respectively. The various components in the domain may include tunable parameters, $\theta$. A domain may also optionally provide a scalar reward $r\in\mathcal{R}\subseteq \mathbb{R}$ signal, where a reward function maps an action and the internal environment state to a scalar value. Note that it is entirely possible that a reward can be provided in the proxy domain but not in the target domain since we don't have direct access to the internal states needed to calculate it. 
Consequently, a domain $S \in \mathcal{S}$ can be considered as something that maps control commands $u \in \mathcal{U}$ to observations $z \in \mathcal{Z}$, conditioned on some environment model state $x \in \mathcal{X}^{env}$ and parameters $\theta \in \Theta$
    ($S:\mathcal{U}\times \mathcal{X}^{env} \times \Theta \mapsto \mathcal{Z} \times \mathcal{R}$). 
We will refer to a single domain generated by a specific set of parameters as a \textbf{domain instance} $S_\theta$ and the set of all domains that are possible to achieve by varying the parameters as a \textbf{domain family}, $S_\Theta$.

A \textbf{task}, $T \in \mathcal{T}$, is specified through one or more \textbf{evaluation metrics}, $M$, which map a trajectory of $N$ states  to a real-valued number:
\begin{equation}
    T\triangleq \{M_i\}_{i=1}^m \text{,}\quad M_i:\mathcal{X}^N \to \mathbb{R}
\end{equation}

An \textbf{agent}\footnote{We distinguish an agent from the more standard notion of \textit{policy} in that a policy maps internal states to actions} contains the algorithm 
that is used to generate control commands from the history of observations, the history of control commands, some initial state $x_0 \in \mathcal{X}^{agent}$ (at time $t$, $A: \mathcal{Z}^{t-1} \times \mathcal{U}^{t-1} \times \mathcal{X}^{agent} \mapsto \mathcal{U}$, or if the state is assumed to be Markovian then $A: \mathcal{Z} \times \mathcal{X}^{agent} \mapsto \mathcal{U}$). Presumably, the algorithm informing this agent is designed to optimize the specified task evaluation metrics. A \textbf{learning agent} \label{def:learning_agent} is able to adapt its behaviour over time by means of a learning algorithm ($A$ at time $k$ is not necessarily the same as $A$ at time $k+1$). 
However, we assume here that, for a stationary domain and task, the learning agent will \textit{converge} to a stationary agent for some $k$ large enough.



Given a sequence $u_{0:n}$ of control values (generated by any means), a domain instance $S_\theta$ can generate a dataset $D_{S_\theta}$, conditioned on an initial state $x_0$, which is a collection of $n$ labelled data samples each consisting of a tuple of observation $z$ and command $u$ ($D_{S_\theta}=\{(z_i,r_i, u_i)\}_{i=1}^n$ where $(z_i,r_i)=S_\theta(x^{env}_i,u_i)$). 
It is also possible that the parameters, $\theta$, are randomized over the domain family during dataset generation. In this case we have $D_{S_\Theta}=\{(z_i,r_i, u_i)\}_{i=1}^n$ where $(z_i,r_i)=S_\theta(x^{env}_i,u_i$) and $\theta \sim \Theta$. 
Such datasets are useful as demonstrations for the learning agent, for example in imitation learning algorithms \cite{ross2011reduction} which try to reproduce the behaviour of an expert.

In an on-policy reinforcement learning setting, the control actions are selected at every timestep: $u_i=A(z_i,x^{agent}_i)$. The domain will update its state according to the command and will return the next observation and optionally a reward to the agent. 
Normally the agent will be able to use the reward to update its algorithm. Again the on-policy rollouts can be executed on the same domain instance or on random samples from the simulator family (as is the case in domain randomization \cite{tobin2017domain}) \footnote{The instances may not be sampled randomly, as is the case in \cite{pmlr-v100-mehta20a}}.

Our objective in this work is to provide measures to quantify the \textit{usefulness} of a proxy domain. We will use the term ``Proxy Usefulness (PU)'' to quantify the value of a proxy domain. 

\subsection{A Naive Measure of Proxy Usefulness}

Naively, one could define the PU of a proxy domain by its discrepancies with the target domain it is expected to represent. Following, the usefulness of a proxy domain would be inversely proportional to its domain gap with the target domain and, according to (Fig.~\ref{fig:agent-environment}), would be defined by:

\begin{definition} [\textbf{\ac{POD}}]
A proxy is deemed more useful inasmuch as the ``difference'' in the resulting observations produced by the proxy's and the target's robots for a predefined sequence of control commands is small.

For a sequence of control values $u_{0..n}$ and initial state, this could simply be calculated as:
\begin{equation}
    || S^{proxy}_\theta(x^{env}_i,u_{i}) - S^{target}_\theta(x^{env}_i,u_{i}) ||
    \label{eq:naive}
\end{equation}
where in this case we are primarily concerned with the observations that are output and not the rewards. 

\label{def:naive_proxy_usefulness}
\end{definition}

There are several issues with Def. \ref{def:naive_proxy_usefulness}:
\begin{enumerate}
    \item It combines in an opaque way the various sources of the discrepancy. Referring to Fig. \ref{fig:agent-environment}, there could be a ``gap'' in the dynamics, the environment model (for example how other agents move in the environment), or in the generation of sensor data based on a rendering model. Moreover, errors in upstream models will compound.  
    \item  It is agnostic to the task that the agent is trying to solve. Many proxies should be deemed perfectly faithful if a trivial task is chosen, but that will not be the case here.
    \item  It presupposes that the fidelity in the target domain is needed. In practice, we only require a form of task-conditioned fidelity: only the things that are important to solve the task at hand must be faithfully reproduced in the proxy. 
    \item It in no way represents how useful the proxy is for learning. 
\end{enumerate}

In the remainder of this paper, we offer alternative ways to evaluate the PU to address these issues.



We make a clear distinction between two different uses of a proxy: 1) The \textit{predictivity} value of a proxy domain (given a target domain) is its ability to generate accurate predictions about the performance of agents in the target domain. 
The \textit{teaching} value of a proxy domain encapsulates how useful it is as a tool to train \textit{learning agents} that perform well on the target domain.

\section{The Proxy as a Predictor}
\label{sec:prediction}

The first ``value'' of a proxy domain is as a tool to predict. However, different from Def.~\ref{def:naive_proxy_usefulness}, we argue that the domain's ability to predict \textit{task performance} rather than exact observations is what is relevant:

\begin{definition} [\textbf{\ac{PPV}}]

Given a task $T$ defined by evaluation metrics $M_{1:m}$, and an agent $A$ that generates trajectory $x^{1:N}_{proxy}$ in the proxy domain and $x^{1:N}_{target}$ in a target domain given equivalent starting conditions,  then we define the \ac{PPV} of a proxy $S_\theta$ as the discrepancy of the resulting evaluation metrics:
\begin{equation}
    \text{PPV}(S_\theta, A) \triangleq \sum_{i=1}^m \beta_i |M_i(x_{proxy}^{1:N}) - M_i(x_{target}^{1:N})|
    \label{eq:PETV}
\end{equation}
\noindent
where the $\beta_i$ terms are weighting constants that can account for mismatched units or possibly increased importance of one metric over another. A proxy is deemed more useful if it has a lower \ac{PPV}.
\label{def:PPU}
\end{definition}

By Def.~\ref{def:PPU}, a proxy domain can be considered \textit{perfectly faithful} to a target domain for a given task if the \ac{PPV} is zero for all possible agents. This definition is in some sense a generalization of the \ac{SRCC} \cite{kadian2019making} to the case where there are multiple metrics that define the task, except with a 1-norm distance instead of the bivariate correlation.  

The need for the $\beta$ constants in Def.~\ref{def:PPU} is undesirable since it allows some room for subjectivity that can effect the results. This can be avoided by considering that in many cases we are interested in \textit{comparing} agents rather than finding exact evaluations of the metrics. As a result, we can consider a relaxation of Def.~\ref{def:PPU} to the relative case. Given that a task may contain several evaluation metrics, agents can be arranged in a partial ordering whose binary relation $\leq$ is defined by dominance along all of the available metrics:
\begin{equation}
    A_1 \geq A_2 \rightarrow M_i(X_1) \geq M_i(X_2) \quad \forall i
\end{equation}

\noindent
where $X_j$ is shorthand for the trajectory produced by agent $A_j$ (either in the proxy or in target domain).  

\begin{definition} [\textbf{\ac{PRPV}}]
Given $K$ agents, the relative predictive ability of a proxy $S_\theta$ is defined by its ability to accurately predict the binary relations between agents that would be present in the target domain. Let $\mathcal{A}^{proxy} = [\alpha^{proxy}_{ij}]_{i,j=1..K}$
be a matrix whose entries are given by:
\begin{equation}
    \alpha^{proxy}_{ij}= \begin{cases}
    1 & A^{proxy}_i \geq A^{proxy}_j \\
    0.5 & A^{proxy}_i \ngeq A^{proxy}_j \quad \& \quad A^{proxy}_j \ngeq A^{proxy}_i \\
    0 & A^{proxy}_i \leq A^{proxy}_j \\
    \end{cases}
\end{equation}
where $A_j^{proxy}$ ($A_i^{proxy}$) is agent $j$ ($i$) applied to the proxy domain. We similarly construct $\mathcal{A}^{target}$. Then the \ac{PRPV} of a proxy domain is given by the 1-norm between the two matrices that represent the relations in the two partial orders \cite{Cook}:
\begin{equation}
    \text{\ac{PRPV}}(S_\theta, A_{1:K}) = \sum_{i,j=1}^K |\alpha_{ij}^{proxy} - \alpha_{ij}^{target}|
    \label{eq:RPPU}
\end{equation}
\label{def:RPPU}
\end{definition}

According to Def.~\ref{def:RPPU}, a proxy domain is now perfectly faithful if it produces the identical partial order over agents that would be produced if the agents were run on the real robot. This is closely related to the concept of ``rank inversion'' \cite{Ligot_swarms}. 

Note that in the case of both \ac{PPV} and \ac{PRPV}, the value of the domain is \textit{conditioned} on the task and \textit{agnostic} to the agent (only requires some method of generating trajectories). Also note that in practice the performance of the agent in a simulated domain or (especially) in the real domain will be stochastic and therefore \ac{PPV} and \ac{PRPV} should be redefined as metrics over distributions and approximated by sequences of trials, but we omit this here for clarity.

\section{The Proxy as a Teacher}
\label{sec:learning}

Orthogonal from the proxy domain's predictivity, it may have value as a tool for agents that \textit{learn} (\ref{def:learning_agent}). That proxy domain now becomes a part of the agent generation process since, as shown by the dashed lines in Fig.~\ref{fig:agent-environment}, the task performance may be fed back to the agent. We can assess the usefulness of the proxy domain by evaluating the performance of the agents that it trains on the target domain, compared to agents that learn entirely on the target domain. A proxy domain is deemed more valuable if it reduces the number of trials that are needed in the target domain.
One naive option to evaluate a domain transfer method would be to consider the zero-shot (or N-shot) performance on the target domain. However, similar to the argument we made in Def.~\ref{def:PPU}, the outputs of these metrics may not calibrate well to the actual learning that has taken place. 
Instead, we define the usefulness for learning explicitly as what we are trying to minimize through using the simulator for training: the number of trials on the target domain required to achieve equivalent performance as we would achieve if we had not pre-trained in the proxy domain. 

\begin{definition} [\textbf{\ac{PLV}}]
    Consider that a learning agent, $A^{target}$ trained entirely in the target domain is able to achieve a performance of $M^{target}_{i..m}$ on task $T$ \textit{at convergence} using dataset $S^{target}$. An agent $A^{proxy\rightarrow target}$ pre-trained on the proxy domain and then transferred to the target domain and fine-tuned until it achieves an equivalent performance $M^{proxy\rightarrow target}_{i..m} \geq M^{target}_{i..m}$ using dataset on the target domain $S^{proxy\rightarrow target}$. Then, the \ac{PLV} is given by:
    \begin{equation}
        \text{\ac{PLV}} = |D^{target}| - |D^{proxy\rightarrow target}|
        \label{eq:LPU}
    \end{equation}
    
    \label{def:LPU}
\end{definition}


Note here that, in contrast to the predictivity, the usefulness of the proxy as a teacher is conditioned on \textit{both the task and the learning algorithm} used to train the agent. 
In addition, different learning algorithms may leverage the proxy domain in different ways.  Behavior cloning methods 
may generate a dataset that includes expert trajectories from both domains \cite{bojarski2016end}. In this case the usefulness is determined by the reduction in the size of the dataset, $D^{target}$ from the target domain (an example we demonstrate in Sec.~\ref{sec:LPU_exp}). As noted in Sec.~\ref{sec:prelim}, the proxy domain dataset can be generated from a single domain instance or over the domain family for increased robustness. In either case, the definition of the usefulness value is unchanged. 

In reinforcement learning, the agent is learning through interaction with the environment \cite{codevilla2018end}. In this case the data is generated through trials and this number of trials is what we seek to minimize. Again, it is entirely possible that the training episodes in the proxy environment randomize over the domain family, as is the case in domain randomization \cite{tobin2017domain}.  

In some cases, both off-policy data and on-policy rollouts may be used. Such is the case in approaches that leverage domain adversarial transfer \cite{zhang2019adversarial, Bharadhwaj_ICRA_2019}. Off-policy data is used to learn the mapping from proxy domain to the source domain (for example using a discriminator). At test time, this mapping is applied but then fine-tuning is achieved with on-policy trials. A generalization of the \ac{PLV} may consider these two cases (on-policy and off-policy) as being scaled differently. A natural priority would be to minimize the number of on-policy trials on the target domain at the expense of off-policy data, which may be easier and safer to obtain. As such we can generalize \eqref{eq:LPU} to:

    \begin{equation}
    \begin{split}
        \text{\ac{PLV}} &= \eta_{on}(|D_{on}^{target}| - |D_{on}^{proxy\rightarrow target}|)  \\
            & + \eta_{off}(|D_{off}^{target}| - |D_{off}^{proxy\rightarrow target}|)
    \end{split}
    \end{equation}
\noindent
where $\eta_{on}$ and $\eta_{off}$ represent the relative importance of off and on-policy data. 


\section{Optimizing the Proxy}

Using any of the proxy domain metrics defined above, we can optimize the set of parameters $\theta$ of the proxy instance. 

In the case of predictivity, we desire to minimize the \ac{PRPV}:
\begin{equation}
\theta^*(A_{1:K}) = \arg \min_\theta \text{\ac{PRPV}}(S_\theta, A_{1:K})
\label{eq:opt_PRPV}
\end{equation}
which will yield the parameters that are most predictive of the task metrics. The residual of this optimization defines the irreducible domain gap and is due to the fact that the proxy models may be limited in their capacity. For example, if wheel slip is not modeled in the proxy domain but it exists in the target domain, we will always incur error. 

In the case of teaching, we desire to maximize the \ac{PLV}. It is possible that we wish to find the single best proxy instance, but more recent methods leverage randomization for increased robustness. In this case, we may wish to optimize for the optimal distribution or sequence of domain instances (e.g., as in \cite{pmlr-v100-mehta20a}), a problem that can be formulated as curriculum learning:
\begin{equation}
    p(\theta)^* = \arg \max_{p(\theta)} \text{\ac{PLV}}(\theta)
\end{equation}

\noindent or possibly over the domain itself:
\begin{equation}
    S^*=\arg \max_S \text{\ac{PLV}}(S)
    \label{eq:opt_proxy_family}
\end{equation}

These optimization may be solved with gradient-based methods in the case that the proxy domain is \textit{differentiable} \cite{murthy2021gradsim, hu2019difftaichi}. Evaluating the loss function may incur significant cost, however, particularly in the case of the \ac{PLV} where training an agent in the proxy domain and evaluating it in the target domain is needed. For these cases, an approach such as Bayesian Optimization or some approximation may be more appropriate \cite{snoek2012practical}. 

\section{Experiments}

We evaluate the proposed metrics in the context of Duckietown \cite{paull2017duckietown}, a platform to run vehicle controllers both in a simulator and on a real robot seamlessly. The platform offers a simulated domain (\cite{gym_duckietown}) where domain randomization is available, as well as multiple customizable features (see Fig. \ref{fig:bothenvs}). A challenge server is in place such that the users can easily submit a Docker container to the server, which will evaluate it and returns footage of the runs as well as performance metrics. Duckietown now additionally offers the evaluation of runs on the real robot at the Autolab  through the challenge server \cite{tani2020integrated}. The Autolab is an embodied domain monitored by watchtowers. As such, users can receive the ground truth position of the robot at each timestamp as well as a precise trajectory of the performance of their agent.


\begin{figure*}[tb]
\centering
\begin{subfigure}[b]{0.24\textwidth}
        \centering
        \includegraphics[width=\columnwidth]{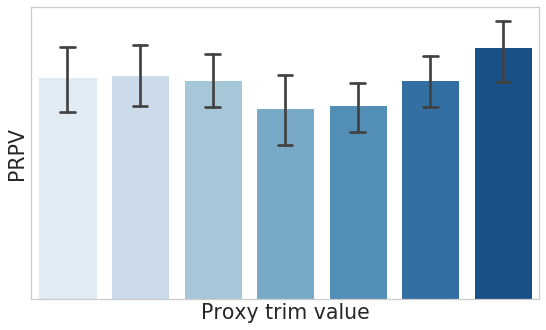}
        \caption{}
        \label{fig:RPPU}
\end{subfigure}
\begin{subfigure}[b]{0.24\textwidth}
         \centering
         \includegraphics[width=\columnwidth]{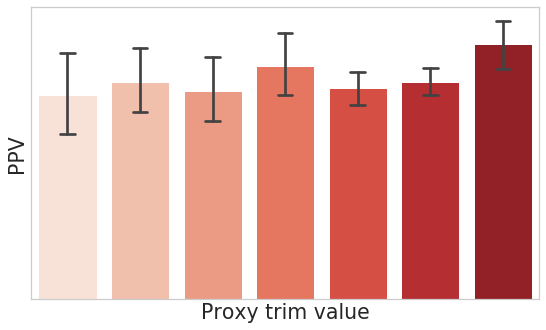}
         \caption{}
         \label{fig:PPV}
\end{subfigure}        
\begin{subfigure}[b]{0.24\textwidth}
         \centering
         \includegraphics[width=\columnwidth]{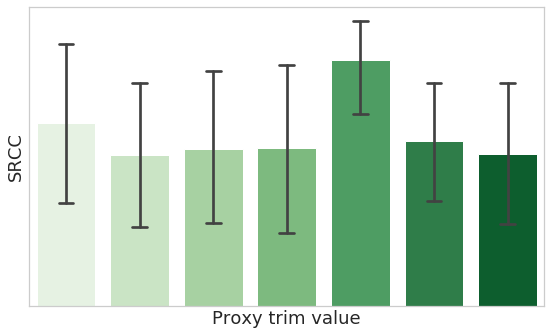}
         \caption{}
         \label{fig:SRCC}
 \end{subfigure}       
\begin{subfigure}[b]{0.24\textwidth}
         \centering
         \includegraphics[width=\columnwidth]{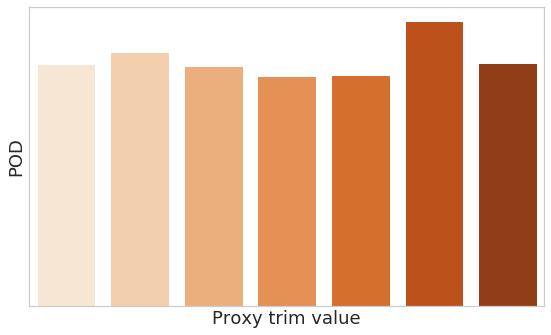}
         \caption{}
         \label{fig:NPV}
\end{subfigure}
\caption{Comparison of proxy usefulness metrics. From left to right: PRPV, PPV, SRCC, POD. For each proxy, the metric is reported for a linear range of trim value from -0.3 to 0.3. The ground truth, 0.0, is always in the center.}
\label{fig:trim}
\end{figure*}

\subsection{Duckietown Simulator Predictivity}
\label{sec:RPPU_exp}

We collected 10 agent submissions from a lane-following challenge from the \ac{AIDO} \cite{aido2019}. Each submission consist of a controller that is designed to drive a Duckiebot along a lane, following the center as closely as possible. The environment configuration in our experiments was a simple 3-by-3 Duckietown map, in which every run would last 60 seconds or until the robot crashes. For a given Duckiebot trajectory, the task is specified through two performance metrics: $M_1$: distance traveled along a lane and $M_2$: survival time, capped at 60 seconds. Since each agent had multiple runs in both the proxy and target domain, the metrics reported are actually means of multiple entries\footnote{The results can be found at \url{https://challenges-stage.duckietown.org/humans/challenges/anc-01-LF-sim-validation/leaderboard}}.

The target domain was set to be the embodied lane following \ac{AIDO} domain, while the proxy domains were instances of the Duckietown simulator \cite{gym_duckietown}. We selected a list of parameters consisting of \textit{trim value}, which controls the wheel trim, \textit{command delay}, which simulates latency in the control loop, \textit{blur time}, which is used to simulate camera blur, and \textit{camera angle}, which is the pitch angle of the camera.  
For some of these parameters (namely the trim value and camera angle) we know the ground truth value on the real Duckiebot. In the case of the trim value, the robot is made to go straight using an odometric calibration procedure, therefore the ground truth value in the simulator should be 0.0. For the camera angle, we can determine it though extrinsic camera calibration.

As a result, we can verify that our metrics are correct since the best predictivity should correspond to the ground truth value. We investigate this for the case of trim in Fig.~\ref{fig:trim}, and compare the scores for \ac{SRCC} (Fig.~\ref{fig:SRCC}) \cite{kadian2020sim2real}, \ac{POD} (Fig.~\ref{fig:NPV}), \ac{PPV} (Fig.~\ref{fig:PPV}) and our proposed \ac{PRPV} (Fig.~\ref{fig:RPPU}). We can see from the plots that the \ac{PRPV} identifies the correct trim value, and also shows monotonic decrease in predictivity as the trim value increases. By writing our metric in terms of agent ordering explicitly we are able to capture the essence of how well the agents are able to perform the task.

\begin{figure}[tb]
\centering
        \centering
        \includegraphics[width=\columnwidth]{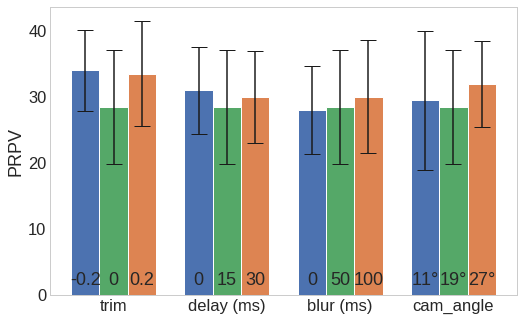}
        \caption{Evaluation of different instances of gym\_duckietown as a proxy for the \ac{AIDO} embodied challenge according to the \ac{PRPV}. The green bars represent the parameter value that was used for \ac{AIDO}. Using our metric we can now determine which of our parameters were correct and which should be modified. }
        \label{fig:prpv}
\end{figure}

We apply a similar approach to the other tunable parameters. 
For each simulator instance, we report the \ac{PRPV} in fig \ref{fig:prpv} for three different parameter values centered on the ones that were actually used in the recent \ac{AIDO} competition. 

As expected, the results in fig. \ref{fig:prpv} show that having a non-zero trim value highly reduces the predictivity of our simulator. 
We also observe that the \textit{command delay} that was initially set to 50 ms seems to be good, same for the \textit{camera angle}. However, in the case of the \textit{blur time}, our simulator could make better predictions if we decrease the parameter value. To obtain the optimal simulator instance, a second iteration of that experience could be repeated, using the new information to choose the range of the parameters to explore.

\subsection{Duckietown Simulator for Learning}
\label{sec:LPU_exp}

\begin{figure}[tb]
\centering
        \centering
        \includegraphics[width=0.9\columnwidth]{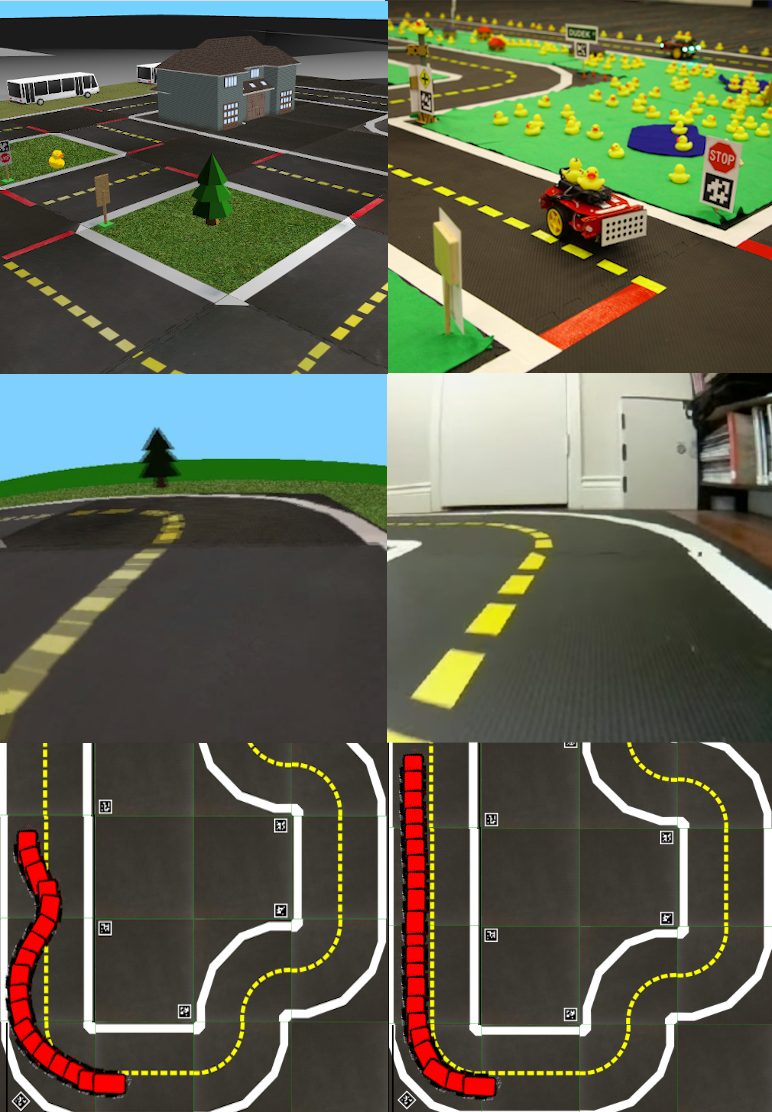}
        \caption{Proxy (left) and target (right) domain of Duckietown, displaying the environment, observation and trajectories.  }
        \label{fig:bothenvs}
\end{figure}

To compute the \ac{PLV}, we collected annotated data from an expert that has access to ground truth both in the proxy and target domains (fig. \ref{fig:bothenvs}). We then used the algorithm of the controller that placed first during AI-DO3 \cite{frankbc}, which is based on imitation learning \cite{bojarski2016end}, to obtain a learning-based agent capable of following a Duckietown lane. Again, the environment configurations was set to be a 3-by-3 Duckietown map, where the agent has to drive within the outer lane for 60 seconds. We note the performance of an agent by the amount of time it can drive, where each second spent with one wheel outside the lane is subtracted and each second spent with both wheels outside the lane is subtracted twice. If the agent drives out of the map entirely, the run ends.

Computing the \ac{PLV} requires comparison of the performance of an agent with and without the proxy domain. We trained the neural network based on various amount of real world samples, each time deploying on the real robot and observing its performance. We observed that the agents converge at a performance of around 55 seconds score (that is, they survive the whole 60 seconds run and get around 5 seconds of penalty), which was obtained with 9000 annotated data points from the target domain. Afterwards, we performed the same experiment, but we initialized the weights of the network from a neural network trained entirely with data from a simulator-based proxy domain.

\begin{table}[tb]
\begin{centering}
\begin{tabular}{ |c| c | c | c| } 
 \hline
 Proxy & zero-shot score & target domain samples & \% reduction\\ 
   \hline
 sim\_ca03 & 4.0 & 1000 & 88\\ 
 sim\_ca19 & 47.4 & 250 & 97\\ 
 sim\_ca35 & 5.4 & 750 & 92 \\ 
 none & 0 & 9000 & 0.0\\ 
 \hline
\end{tabular}
\caption{Zero shot score and number of target domain samples required to achieve real world performance at convergence on the robot according to proxy used for pre-training.}
\label{table:1}
\end{centering}
\end{table}

\begin{figure}[tb]
\centering
        \centering
        \includegraphics[width=\columnwidth]{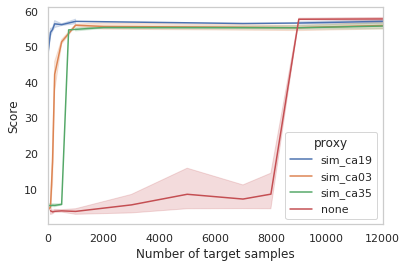}
        \caption{Performance of an imitation learning agent in a 3x3 duckietown map lane-following challenge with different proxy domains}
        \label{fig:lpu}
\end{figure}

The experiment was done for three different proxy instances where we modified the angle of the simulated camera to 3°, 19° and 35°, respectively. The learning curves for the four cases are shown in  Fig. \ref{fig:lpu} and results are presented in Table \ref{table:1}.
We see here that the zero-shot score on the target domain is not a representative value for how useful the simulator is for learning. The zero-shot scores for `sim\_ca03' and `sim\_ca35' are very low compared to `sim\_ca19'. However, they are both still very useful since they dramatically reduce the amount of target domain data needed compared to training entirely on the target domain. 
We also see here that these two different uses of a simulator (predictivity and teaching) may tell us different things. A simulator that is not particularly predictive (since the camera pitch angle is way off) may still be useful for learning if the learning algorithm is able to adapt properly.

\section{Conclusion and Future Work}

We introduce new  metrics to assess the usefulness of proxy domains for agent learning. In a robotics setting it is common to use simulators for development and evaluation to reduce the need to deploy on real hardware.
We argue that it is necessary to  to take into account the specific task when evaluating the usefulness of the the proxy. We establish novel metrics for two specific uses of a proxy. When the proxy domain is used to predict performance in the target domain, we offer the \ac{PRPV} to assess the usefulness of the proxy as a predictor, and we argue that the task needs to be imposed but not the agent. When a proxy is used to generate training data for a learning algorithm, we propose the \ac{PLV} as a metric to assess usefulness of the source domain, which is dependent on a specific task and a learning algorithm.
We demonstrated the use of these measures for predicting parameters in the Duckietown environment. Future work will involve more rigorous treatment of the optimization problems posed to find optimal parameters, possibly in connection with differentiable simulation environments. 

\section*{Acknowledgements}

Liam Paull is supported by the Canada CIFAR AI Chairs Program. The work was also supported by the National Science and Engineering Research Council of Canada under the Discovery Grant Program and by Samsung Electronics Co. Ldt. All of which we gratefully acknowledge. The authors would also like to thank Chude (Frank) Qian for his contributions to the imitation learning pipeline and Charlie Gauthier for additional support.  





\bibliographystyle{IEEEtran}
\bibliography{bib}


\end{document}